# DEVELOPMENT OF A FIRE DETECTION SYSTEM ON SATELLITE IMAGES[1]

2022g. A.N. Averkin[a] S.A. Yarushev[a]

[a]Plekhanov Russian University of Economics, Moscow, Russia
e-mail: averkin2003@inbox.ru, sergey.yarushev@icloud.com

This paper discusses the development of a convolutional architecture of a deep neural network for the recognition of wildfires on satellite images. Based on the results of image classification, a fuzzy cognitive map of the analysis of the macroeconomic situation was built. The paper also considers the prospect of using hybrid cognitive models for forecasting macroeconomic indicators based on fuzzy cognitive maps using data on recognized wildfires on satellite images.

**Architecture of the developed deep neural network**

It is necessary to dwell in more detail on the key points of the architecture of the neural network being developed. The loss function demonstrates how accurately the developed machine learning model is able to predict the expected result. In our case, binary cross entropy is used as a loss function. Cross-entropy shows the difference between two probability distributions. When the cross entropy is large, it means that the difference between the two distributions is large, and if the cross entropy is small, then the distributions are similar to each other. In the case of binary classification and the application of binary cross-entropy, each predicted probability is compared with the actual value of the class (0 or 1), thus calculating an estimate that fines the probability based on the distance from the expected value. Figure 1 shows the range of possible values of the binary cross-entropy function when accounting for the truth of the observation ($y = 1$). When the predicted probability approaches 1, the loss function gradually and slowly decreases. But, however, if you reduce the predicted probability, it begins to increase very quickly [1].

---

[1] This research was performed in the framework of the state task in the field of scientific activity of the Ministry of Science and Higher Education of the Russian Federation, project "Development of the methodology and a software platform for the construction of digital twins, intellectual analysis and forecast of complex economic systems", grant no. FSSW-2020-0008

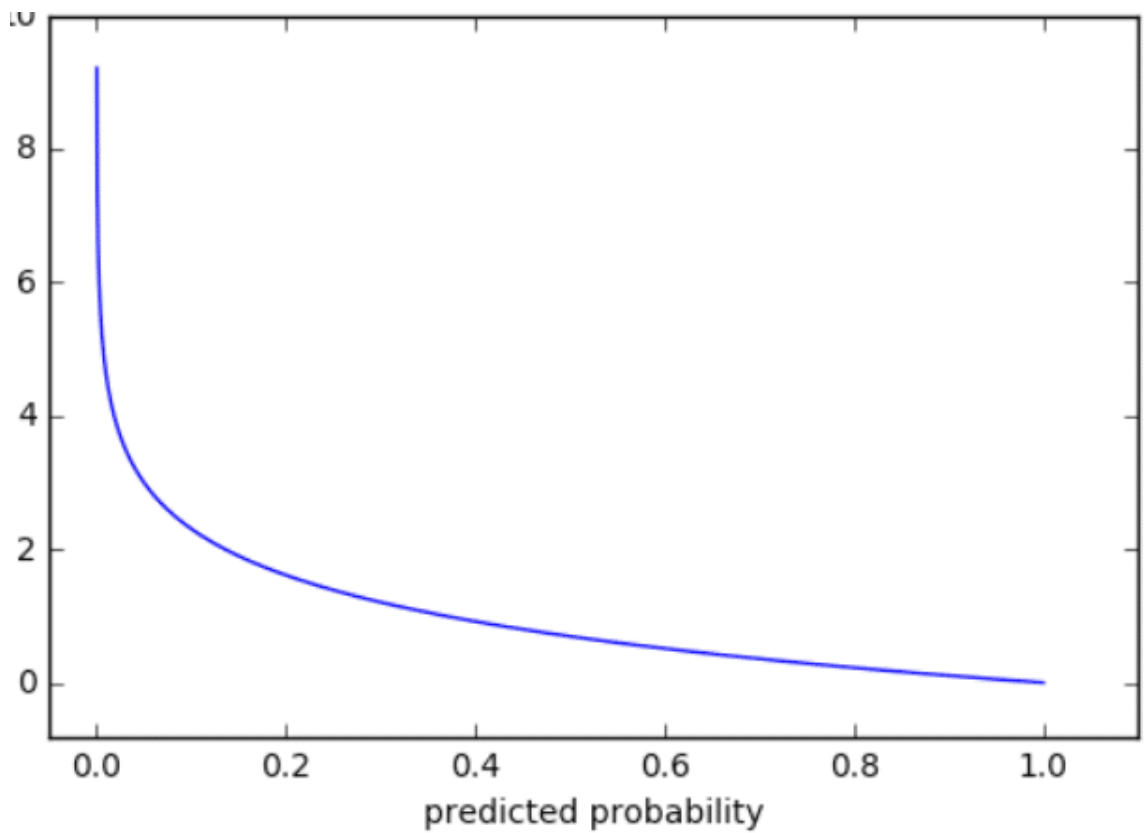

Figure 1 – Visualization of the binary cross-entropy function at the truth of the observation

Figure 2 shows an example of neuron Y.

$$Y = \sum (weight * input) + bias$$

Figure 2 – Example of neuron Y

The value of the neuron can be located in the range from -∞ to + ∞. In fact, the neuron does not know the boundary after which activation should occur. In order to decide whether a neuron should be activated just and an activation function is added. It checks the value of a neuron to see if external connections need to treat the neuron as activated or ignore it.

In the developed convolutional neural network, 2 activation functions are used on different layers - sigmoidal and ReLu.

The sigmoidal function is nonlinear in itself, and a combination of several sigmoidal functions will also produce a nonlinear function. It is also worth paying attention to the fact that the sigmoid is non-binary, which makes the activation process analog.

With a change in x between -2 and +2, Y changes at an incredible rate, which only means that even with the most insignificant change in X in a given range, Y will change dramatically.

The sigmoidal function is great for being used in classification-related tasks. It tries to bring values to one of the edges of the curve, this behavior makes it possible to find specific boundaries when predicting. Another advantage of the sigmoidal function over, for example, the linear function is the following fact: the sigmoidal function has a fixed range of values (from 0 to 1), while the linear function varies from -∞ to +∞. This property of the sigmoidal function is very useful, because it does not lead to errors at large activation values. That is why the sigmoidal function today is one of the most popular activation functions when solving problems using neural networks.

The ReLu activation function is $A(x) = \max(0,x)$, that is, it becomes clear from the definition that ReLu returns x if x is a positive number, and 0 if x is a negative number.

ReLu is a nonlinear function, and the combination of several ReLu is also nonlinear. The range of ReLu values ranges from 0 to ∞. A big advantage of the ReLu function is the rarefaction of activation, because unlike, for example, a sigmoidal function or hyperbolic tangent, ReLu can be used in complex deep networks with a large number of neurons. This is possible because ReLu it is much less computational power-intensive than sigmoid and hyperbolic tangent, since ReLu performs much simpler mathematical operations. That is why it is appropriate to use ReLu when creating deep neural networks [2].

The method of learning – the method of backpropagation of the error is probably the most fundamental part of the neural network. The meaning of the method of reverse propagation of error is to determine the weight of each neuron in exactly the proportion in which this neuron affects the overall error rate. If you gradually reduce the weight error of each neuron, then in the end you will get such a set of weights that will make, indeed, excellent predictions. Thus, the backpropagation method aims to minimize the function of loss by adjusting the weights and displacements of the network.

The training sample consisted of 2,000 images. The loss value for the training sample was 0.2154, and the accuracy value was 0.9135. For the test sample, respectively, the loss value was 0.0756, the accuracy was 0.9728.

**Analyzing Training Data Sources**

As a source of data, namely satellite images, 2 sets of data will act: this is the Chinese dataset of satellite images FAIR1M and an extensive collection of satellite images from the American company DigitalGlobe.

FAIR1M is a huge dataset of high-resolution satellite images designed for machine learning tasks, namely the detailed recognition of objects in remote sensing images, that is, satellite images. The dataset contains more than 40,000 images, with each image tagged into 5 different categories and 37 sub-categories.

DigitalGlobe is an American company that is the operator of civil satellites for remote sensing of the Earth. The company has at its disposal 5 civilian satellites that take ultra-high resolution images of the Earth. Access to the images is carried out using the ArcGIS Living Atlas of the World application [3]. The World Imagery collection allows you to look at aerospace photographs and satellite images of the planet with an accuracy of 1 meter. A distinctive feature of the collection is the ability to view archival images of the planet (up to 100 different versions at different intervals).

Let's move on to the architecture of the developed convolutional neural network, which consists of 5 layers:

- convolutional layer;
- Pooling layer (subdiscretizing layer);
- anti-aliasing layer with no parameters;
- 2 fully connected neural layers.

It is worth noting that the network does not have an input layer as such, because the dimension of the input tensor is determined in the first layer (in our case, in the convolutional layer). On fully connected layers, the activation functions relu and sigmoidal (sigm) are used. Fully connected neural layers consist of 128 neurons and 1 neuron, respectively. Binary cross entropy is used as the loss function, and accuracy is used as a metric. The learning method is the method of reverse propagation of the error.

**The results of the recognition of wildfires based on the developed neural network**

Let's move on to the results of the developed model. Figures 3,4 show satellite imagery that was classified by the model as a fire and assigned to the "fire" class. To localize the fire in the case of classification of the image as "fire", the myplotlib library is used. The area of the fire is calculated as the area of the rectangle, which is obtained by the coordinates of the extreme of the brightest pixels on each of the axes.

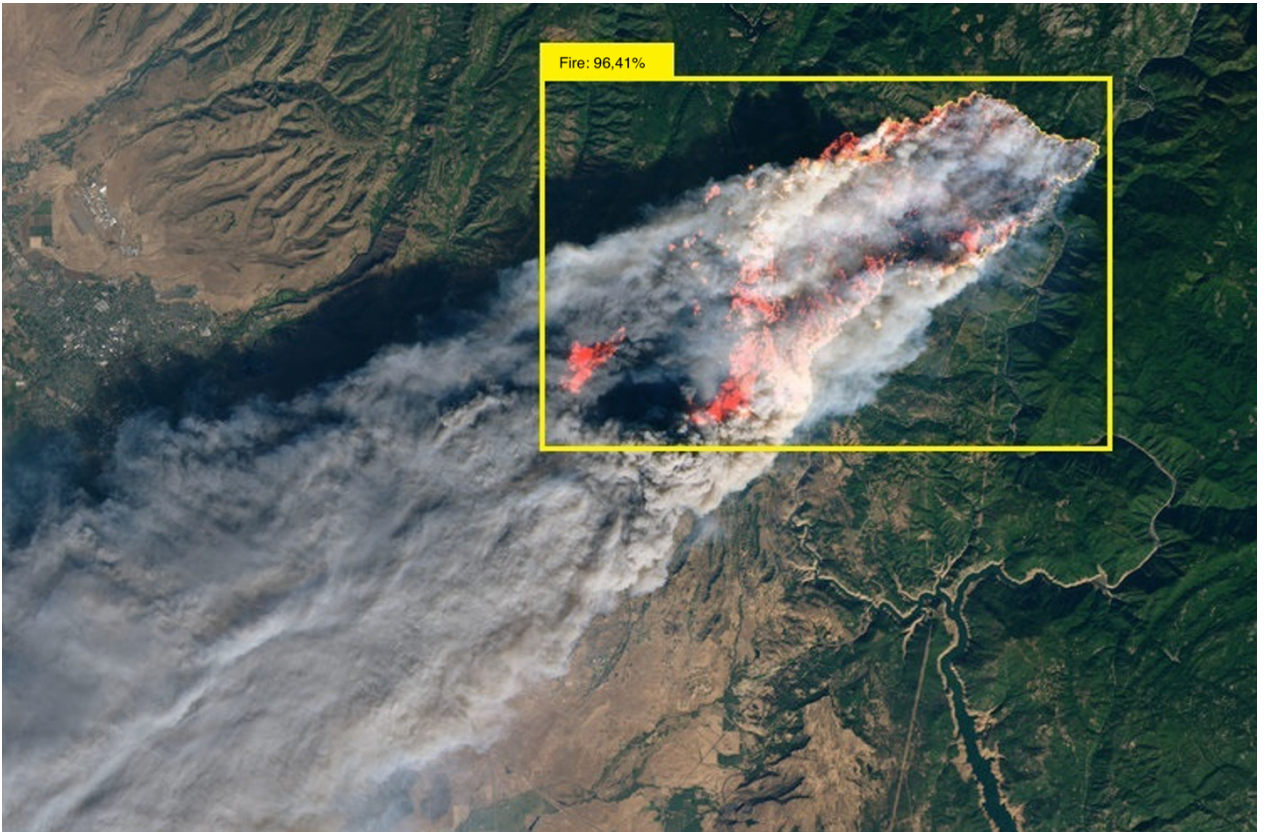

Figure 3 – Image classified by the network as "fire"

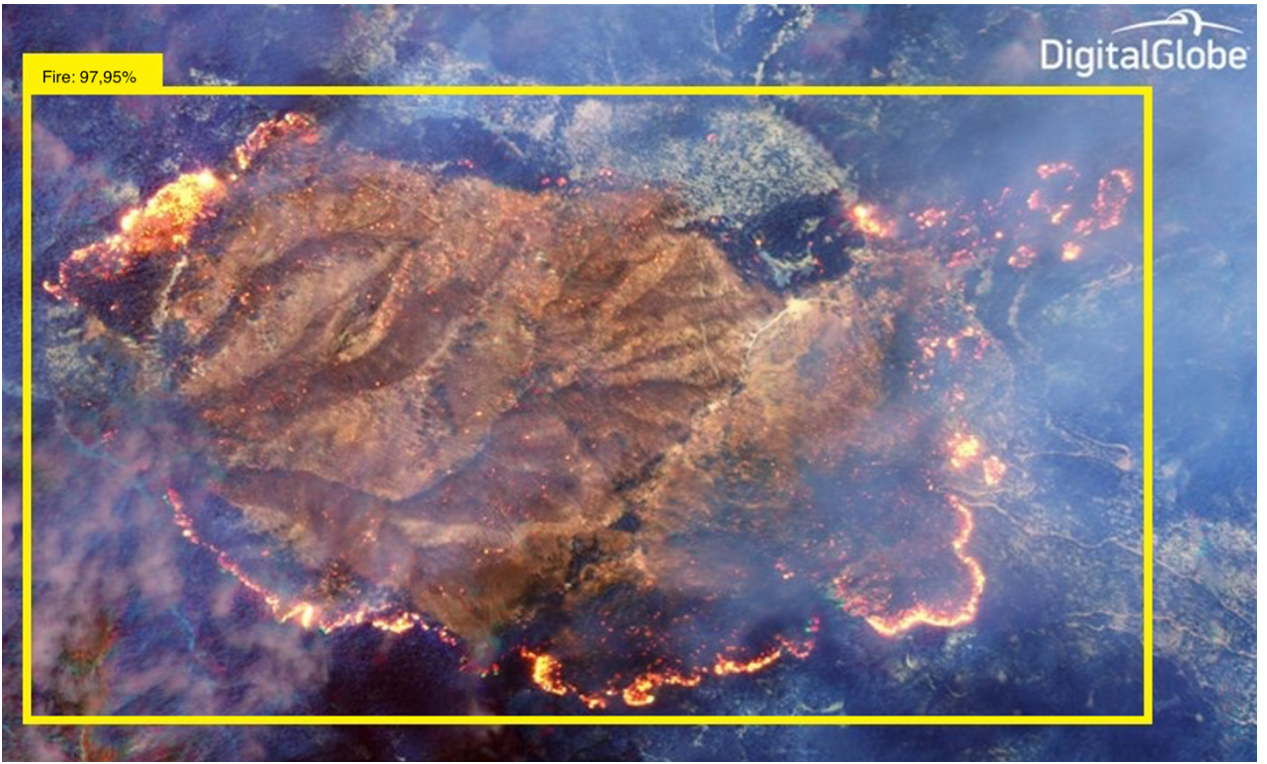

Figure 4 – Image classified by the network as "fire"

# Using a model based on a fuzzy cognitive map, taking into account fire detection in a satellite image

If we automate the process of recognizing wildfires, it will not be difficult to conduct operational analytical studies on the frequency of natural disasters, their scale and seasonality in an automated mode. Also, as a prospect, it is planned to supplement the fire recognition system with the ability to automatically assess the scale of fires and predict their possible direction of spread. These parameters can be used in the future to assess the impact on the macroeconomy of the affected regions.

To use the information obtained from the classification of satellite images in order to predict the impact of natural disasters on the macroeconomy, fuzzy cognitive maps (NCC) are best suited. The NCC model was developed by the authors of the study and published in previous studies [4]. To understand the work of the NCC, dadim formal definition of a fuzzy cognitive map: A fuzzy cognitive map is an orgraph, the vertices of which are concepts or key factors in the development of the situation, and arcs are cause-and-effect relationships between them.

Let $G = \{E, W\}$, E be the set of concepts, W be the set of connections (adjacency matrix): w: : . For example, a primitive cognitive map of the sanitary condition can be specified by the following concepts: e1 – the population of the city; e2 – migration of the population to the city; e3 – the level of modernization of production; e4 – number of urban landfills; e5 is $E \times E \to [-1, 1] e_i, e_j \in E, w(e_i, e_j) \in W$ the frequency of wildfires; e6 is diseases per thousand people; e7 is the prevalence of bacteria in the environment.

The set of connections in a fuzzy cognitive map is conveniently represented as a weighted matrix of contiguity, which we will call a cognitive matrix (in accordance with the terminology adopted by Power [5]). The cognitive matrix for the problem of sanitary condition is given in Table 1.

Table 1. Cognitive matrix "The problem of sanitary condition"

|   | *1* | *2* | *3* | *4* | *5* | *6* | *7* |
|---|---|---|---|---|---|---|---|
| 1 | 0 | 0 | .6 | .9 | 0 | 0 | 0 |
| 2 | 1 | 0 | 0 | 0 | 0 | 0 | 0 |
| 3 | 0 | .7 | 0 | 0 | .9 | 0 | 0 |
| 4 | 0 | 0 | 0 | 0 | 0 | 0 | .9 |
| 5 | 0 | 0 | 0 | 0 | 0 | -.9 | .9 |
| 6 | -.3 | 0 | 0 | 0 | 0 | 0 | 0 |
| 7 | 0 | 0 | 0 | 0 | 0 | .8 | 0 |

Each link is specified with a value between -1 and 1. Instead of numbers, it is convenient to use linguistic scales. In this case, the Expert Advisor uses the following steps to determine the relationships in the cognitive map:

1. Determines which concepts have a cause-and-effect relationship.
2. Defines a linguistic scale consisting of a set of fuzzy linguistic values (or terms), for example: "extremely weak", "moderately", "stronger than usual", etc.
3. For each linguistic therm, it determines the corresponding numerical value in the interval [-1, 1].
4. Specifies the strength of connections using linguistic terms.

In the classical case, connections of a type (the main diagonal of the adjacency matrix) are assumed to be equal to zero, but there are approaches that allow expanding the functionality of a fuzzy cognitive map in the case of such "loops" [$w_{ii}$6].

*The function* of activity of the concepts of the system is defined as $C: E_i \rightarrow C_i$. Each node is mapped to a measure of activity at time *t*. It can take values from 0 (no activity) to 1 (active). *C(0)* specifies the vector of the initial values of node activity. *C(t)* is the vector of states (activity) of nodes in iteration *t*.

**Conclusion**

In this paper, a brief overview of training datasets with satellite images suitable for training an artificial neural network was carried out. The results of the recognition of wildfires on satellite images were demonstrated. As a result of the research, the authors came to the conclusion of the prospects for using automated satellite image recognition systems for the further use of the results in forecasting macroeconomic indicators. In particular, the architecture of a fuzzy cognitive map and an example of the use of fire data in the development and construction of a forecast based on a fuzzy cognitive map are considered. In the future, the authors plan to supplement the functionality of the satellite image recognition model with analytical tools, such as automatic calculation of the fire area and forecasting the possible direction of fire spread. It is worth noting that the recognition of wildfires is only an example of using the potential of deep neural networks to recognize satellite images, since the neural network can be easily trained to solve the problem of recognizing any natural disasters and determining their scale.